\documentclass{article}
\usepackage{arxiv}
\usepackage[utf8]{inputenc}
\usepackage{graphicx}
\graphicspath{ {./images/} }
\usepackage{booktabs}
\usepackage{hyperref}

\title{TrimBERT: Tailoring BERT for Trade-offs}

\author{Sharath Nittur Sridhar \\
    \small{Intel Labs, Intel Corporation} \\
    \small{sharath.nittur.sridhar@intel.com} \\\And
    Anthony Sarah \\
    \small{Intel Labs, Intel Corporation} \\
    \small{anthony.sarah@intel.com} \\\And
    Sairam Sundaresan \\
    \small{Intel Labs, Intel Corporation} \\
    \small{sairam.sundaresan@intel.com} \\
}

\date{}

\begin{document}

\maketitle

\begin{abstract}
Models based on BERT have been extremely successful in solving a variety of natural language processing (NLP) tasks. Unfortunately, many of these large models require a great deal of computational resources and/or time for pre-training and fine-tuning which limits wider adoptability. While self-attention layers have been well-studied, a strong justification for inclusion of the intermediate layers which follow them remains missing in the literature. In this work, we show that reducing the number of intermediate layers in BERT\textsubscript{BASE} results in minimal fine-tuning accuracy loss of downstream tasks while significantly decreasing model size and training time.  We further mitigate two key bottlenecks, by replacing \textit{all} \texttt{softmax} operations in the self-attention layers with a computationally simpler alternative and removing half of all \texttt{layernorm} operations. This further decreases the training time while maintaining a high level of fine-tuning accuracy.
\end{abstract}

\section{Introduction}

Language model pre-training has led to a number of breakthroughs in NLP \cite{devlin2018bert}, \cite{peters2018deep}, \cite{liu2019roberta}, \cite{brown2020language} and achieved state-of-the-art results on many non-trivial NLP tasks. As the accuracy of the BERT model \cite{devlin2018bert} model and its derivatives has increased, so have their size, growing from 110 million (BERT\textsubscript{BASE}) to 175 billion (GPT-3 \cite{brown2020language}) parameters. This makes the already time-consuming task of training unfeasible for many practitioners. In this work, we propose to alleviate this by removing blocks of intermediate layers from BERT-based models which significantly decreases model size and training time.

In addition, certain operations such as \texttt{softmax} are well-known bottlenecks for training BERT. We adapt the work of \cite{richter2020normalized} and apply it to BERT-based models along with other architectural changes to replace all \texttt{softmax} operations. We also show that half of all \texttt{layernorm} operations can be removed from the BERT model with a negligible change in fine-tuning accuracy.

\section{Related Work}

\subsection{Layer Removal}

To decrease the number of parameters, approaches such as ALBERT \cite{lan2020albert} employ a number of techniques, such as decomposing the vocabulary matrix and cross-layer parameter sharing, to achieve similar accuracy to BERT but with 18x fewer parameters and 1.7x faster training. The authors of \cite{press2020improving} changed the ordering of the sub-layers to create the \textit{sandwich transformer}. Like us, the authors recognize that the ordering of the sub-layers in BERT networks is not well justified nor necessarily optimal. However, unlike ALBERT, their goal is to improve language modeling performance and not to decrease the network size or complexity. In this work, we share similar goals to ALBERT but employ a different approach that removes blocks of intermediate layers from BERT to decrease model size and training time.

\subsection{Softmax Replacement}

The bottleneck problem caused by \texttt{softmax} is well-known and was addressed in \cite{stevens2021softermax} by proposing \textit{Softermax} which includes techniques such as changing the exponential function used in \texttt{softmax} from $e^x$ to a low-precision implementation of $2^x$ that results in 2.35x the energy efficiency at 0.90x the size. However, this requires specialized hardware since typical hardware does not support low-precision functions such as exponentiation and division. Our solution does not require specialized hardware and is adapted from \cite{richter2020normalized} to replace all \texttt{softmax} operations with simple normalization resulting in further decreases to the training time of BERT.

\subsection{LayerNorm Removal}

The need for \texttt{layernorm} operations is based on empirical results showing that it, and other normalization techniques, solve the exploding and vanishing gradient problem while accelerating convergence. The authors of \cite{zhang2019fixup, pmlr-v119-huang20f} propose techniques to solve this problem by initializing parameters of the network in such a way as to preclude the need for \texttt{layernorm} operations in Transformer architectures. In our work, we show that half of all \texttt{layernorm} operations can be removed from BERT resulting in an increase in accuracy.

\section{Methods}

\subsection{Layer Removal}

\begin{figure}[h]
    \center{\includegraphics[width=0.7\linewidth] {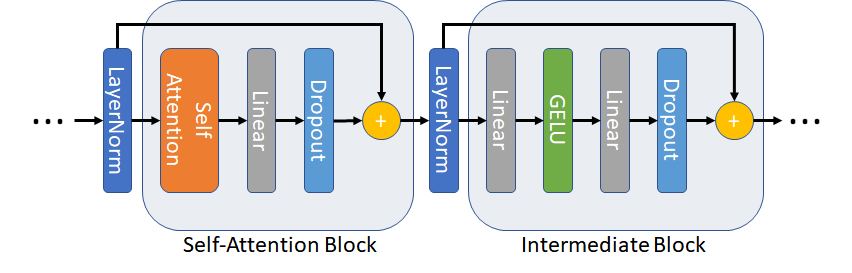}}
    \caption{\label{fig:reference_unit}Reference unit in the unmodified BERT network. The unmodified BERT\textsubscript{BASE} network contains 12 of these units arranged sequentially.}
\end{figure}

\begin{figure}[h]
    \center{\includegraphics[width=1.0\linewidth] {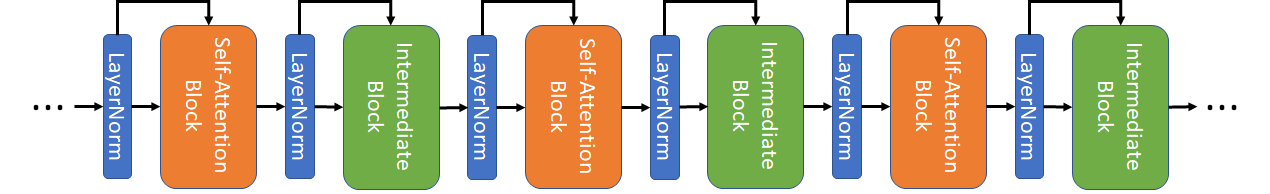}}
    \caption{\label{fig:unmodified_network}First six blocks of the unmodified network architecture. This BERT\textsubscript{BASE} network will have a total of 12 self-attention blocks and 12 intermediate blocks.}
\end{figure}

BERT \cite{devlin2018bert} is a stack of multiple Transformer encoder blocks with each encoder block further consisting of a separate multi-head self-attention block followed by an intermediate block (see Figure \ref{fig:unmodified_network}). The importance of the self-attention block has been extensively analyzed in \cite{clark2019does}, \cite{michel2019sixteen}, \cite{tenney2019bert} and others. An intermediate block is a four-layer feed-forward network containing a Gaussian error linear unit (GELU) \cite{hendrycks2020gaussian} in between two linear layers followed by a dropout layer (see \textit{Intermediate Block} of Figure \ref{fig:reference_unit}). The intermediate block was added mainly to enrich the representations obtained from the self-attention block. However, the relevance of this block has not been well studied and no strong justification for their inclusion in BERT-based networks has been made in the literature.

\begin{figure}
    \center{\includegraphics[width=1.0\linewidth] {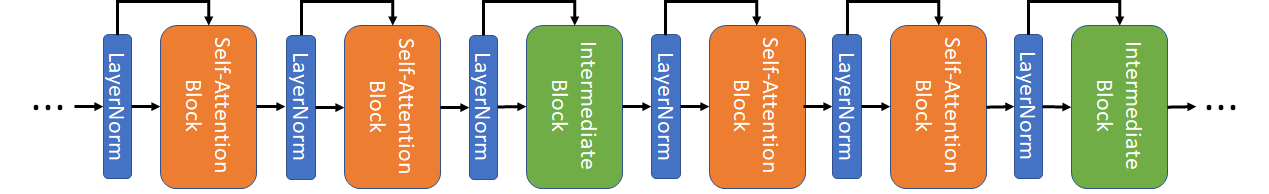}}
    \caption{\label{fig:modified_network}First six blocks of the modified network architecture with $n$ = 2. Note the removal of intermediate blocks between every $n$ = 2 self-attention blocks (compare to Figure \ref{fig:unmodified_network}).}
\end{figure}

Our motivation is to decrease model size and complexity while quantifying any negative impact of reducing the number of intermediate blocks on model accuracy. To this end, we modify the BERT\textsubscript{BASE} architecture by removing some of these blocks within the network. More specifically, an intermediate block will be added only after every $n$ self-attention blocks. If the total number of self-attention blocks in the network is $m$ then the modified network will contain $\lfloor\frac{m}{n}\rfloor$ intermediate blocks. For example, when $n$ = 1 the network is unmodified and contains $m$ intermediate blocks. However, when $n$ = $\infty$ the modified network contains no intermediate blocks. See Figure \ref{fig:modified_network} for an example of the modified BERT\textsubscript{BASE} network with $n$ = 2. Note that $n$ is an architectural hyper-parameter that can be changed to make trade-offs in network size, complexity and accuracy. We experiment with different values of $n$ (see Table \ref{tab:glue_tasks}) and analyze its effects with multiple fine-tuning tasks. 

\subsection{Softmax Replacement}

\begin{figure}[h]
    \center{\includegraphics[width=0.8\linewidth] {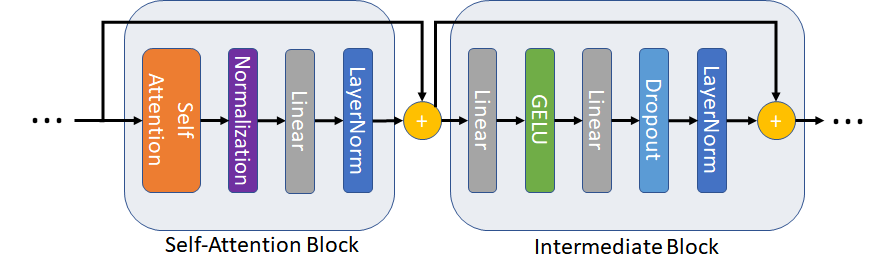}}
    \caption{\label{fig:bandd_reference}Reference unit used in the BANDD BERT\textsubscript{BASE} modified network (compare to Figure \ref{fig:reference_unit}).}
\end{figure}

The authors of \cite{richter2020normalized} propose replacing the \texttt{softmax} operation with a simple normalization of the query-key dot-product when computing self-attention scores. Additionally, the authors introduce certain architectural changes to BERT which include moving the \texttt{layernorm} before the residual connection, adding additional \texttt{layernorm} operations and GELU activation layers as well as other training parameter changes. The experiments in \cite{richter2020normalized} were mainly conducted on synthetic tasks with the goal of being able to better understand the learning dynamics across multiple tasks. In our work, we extend this normalization approach to language model pre-training and subsequent fine-tuning on downstream tasks. Although we use the same normalization technique to replace \texttt{softmax}, we modify the BERT network differently. We identified that removing the dropout layer after self-attention is critical for convergence and accuracy of the modified network. To compensate for the removal of dropout, we instead use L2 normalization in the loss function during pre-training. We call our modifications \textit{BERT Attention Normalization with Divided Dropout} or BANDD (see Figure \ref{fig:bandd_reference} for the reference unit). 

\subsection{LayerNorm Removal}
\begin{figure}[h]
    \center{\includegraphics[width=1.0\linewidth] {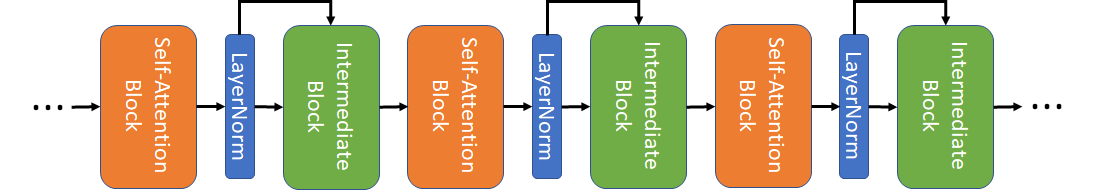}}
    \caption{\label{fig:nomlpln_reference}First six blocks of the NoMLPLN network architecture (compare to Figure \ref{fig:unmodified_network}).}
\end{figure}

Since its introduction, \texttt{layernorm} \cite{ba2016layer} has been an integral part of Transformer-based models because it facilitates faster training, smoother gradients and better generalization accuracy through normalization of intermediate layer distributions. However, it is not without limitations. In \cite{DBLP:journals/corr/abs-1911-07013}, the authors show that the gain and bias terms of each \texttt{layernorm} can increase the risk of overfitting as they are learned from the training set and therefore do not adapt to the test set distribution. Inspired by this insight, we evaluated the performance of BERT\textsubscript{BASE} after removing \texttt{layernorm} operations in each block. While \cite{pmlr-v119-huang20f} showed that it is possible to remove \texttt{layernorm} operations altogether by changing the weight initialization in Transformers, their approach does not transfer to BERT\textsubscript{BASE} resulting in a steep drop in accuracy. Just removing \texttt{layernorm} operations that occur after each self-attention block results in the training process diverging. However, removing \texttt{layernorm} operations that occur after each intermediate block (see Figure \ref{fig:nomlpln_reference}) results in increased fine-tuning accuracy. This suggests that these \texttt{layernorm} operations are causing the model to overfit. We call this modified BERT\textsubscript{BASE} architecture \textit{No MLP LayerNorm} (NoMLPLN).

\section{Results}

In this section we present the accuracy and throughput results for all proposed modifications to BERT\textsubscript{BASE}, which include layer removal, BANDD and NoMLPLN. We combine multiple modifications to create different BERT\textsubscript{BASE} variants and study the effect on accuracy and throughput.  All experiments are performed using BERT\textsubscript{BASE} due to the extremely long times required to train larger models. 

\subsection{Pre-Training} \label{section:pre_training}

All variants were pre-trained from scratch on the English Wikipedia and BookCorpus (\cite{zhu2015aligning}) data sets. To provide a fair comparison between unmodified and modified BERT\textsubscript{BASE} networks, we use the same hyper-parameter values for all model variants. Specifically, we use a sequence length of 128 for 900K iterations with a batch size of 256. An initial learning rate of $10^{-4}$ was used including a linear warm-up schedule for the first 10K steps of training.

\begin{table*}[!ht]
    \resizebox{\textwidth}{!} {%
    \centering
    \small
    \begin{tabular}{@{}lccccccccc@{}}
    \toprule
    \bf Network & \bf MNLI-m/mm & \bf QNLI & \bf QQP & \bf SST-2 &
    \begin{tabular}{@{}c@{}} \bf Parameter \\ \bf Count\end{tabular} &
    \begin{tabular}{@{}c@{}} \bf Size \\ \bf Decrease\end{tabular} & \begin{tabular}{@{}c@{}} \bf Throughput \\ \bf Increase GPU\end{tabular} & 
    \begin{tabular}{@{}c@{}} \bf Throughput \\ \bf Increase CPU \end{tabular}  \\
    \midrule 
    $n$ = 1 (BERT\textsubscript{BASE})     & 82.96/83.27 & 90.01 & 87.45 & 91.62  & 110M & 1.00x & 1.00x & 1.00x \\
    $n$ = 2     & 81.89/82.35 & 89.60 & 87.30 & 89.33 & 81.76M & 1.35x & 1.27x & 1.15x \\
    $n$ = 3     & 81.27/82.36 & 89.14 & 86.97 & 90.13 & 72.31M & 1.52x & 1.39x & 1.21x \\
    $n$ = 4     & 81.20/81.91 & 89.35 & 86.77 & 90.36 & 67.59M & 1.63x & 1.45x & 1.23x \\
    $n$ = 6     & 80.81/80.96 & 88.89 & 86.41 & 89.00 & 62.86M & 1.75x & 1.53x & 1.28x  \\
    $n$ = $\infty$  & 79.19/79.42 & 87.20 & 85.40 & 89.00 & 53.41M & 2.06x & 1.72x & 1.38x \\
    \textit{BANDD} & 80.99/81.43 & 89.12 & 87.68 & 91.05 & 110M & 1.00x & 1.03x & 1.04x  \\
    \textit{NoMLPLN} & 82.66/83.43 & 90.37 & 90.88 & 91.39 & 110M & 1.00x & 1.04x & 1.06x  \\
    \textit{BANDD + NoMLPLN}& 80.90/80.69 & 88.98 & 87.17 & 90.13 & 110M & 1.00x & 1.05x & 1.07x \\
    \textit{BANDD + $n$ = 2}& 80.35/80.46 & 87.37 & 86.98 & 89.68 & 81.76M & 1.35x & 1.34x & 1.19x  \\
    \textit{NoMLPLN + $n$ = 2} & 81.32/81.48 & 89.73 & 90.35 & 90.94 & 81.76M & 1.35x & 1.31x & 1.21x \\
    \textit{BANDD + NoMLPLN + $n$ = 2}& 79.10/78.88 & 86.89 & 85.67 & 89.22 & 81.76M & 1.35x & 1.34x & 1.22x\\
    \textit{BANDD + NoMLPLN + $n$ = $\infty$}& 76.70/77.15 & 85.17 & 85.54 & 87.38 & 53.41M & 2.06x & 1.85x & 1.42x\\
    \textit{DistilBERT}& 79.0/- & 85.3 & - & 90.7 & 66M & 1.66x & 1.81x & 1.89x \\
    \textit{BERT-PKD}& 81.3/- & 88.4 & - & 91.3 & 66M & 1.66x & 1.81x & 1.89x  \\
    \textit{BERT-of-Theseus}& 82.3/- & 89.5 & - & 91.5 & 66M & 1.66x & 1.81x & 1.89x  \\
    \bottomrule
    \end{tabular}}
    \caption{Fine-tuning results for unmodified and modified BERT\textsubscript{BASE} networks along with distillation techniques from \cite{xu2020bertoftheseus}. The m/mm scores for MNLI, F1 score for QQP and accuracy for SST-2 and QNLI on the respective development sets. Also shown is the relative decrease in size and increase in inference throughput with batch size 32 and sequence length 128. The CPU used was an Intel Xeon 8280 and the GPU was a NVIDIA A100.}
    \label{tab:glue_tasks}
\end{table*}

\subsection{Fine-Tuning} \label{section:fine_tuning}

During fine-tuning, all weights of each network investigated are modified through training with a task-specific data set. We evaluate our models on the The General Language Understanding Evaluation (GLUE) benchmark \cite{wang2019glue}. We chose MNLI, QNLI, QQP and SST-2 as our primary tasks since they have sufficiently large data sets and provide stable results across different trials. 

From Table \ref{tab:glue_tasks} we see that the modified BERT\textsubscript{BASE} networks after layer removal perform quite well compared to the baseline. Specifically, with $n \in \{2, 3, 4, 6\}$ there is less than 2\% loss in accuracy for most fine-tuning tasks while simultaneously providing a significant improvement in size and throughput (see Table \ref{tab:glue_tasks}). These results indicate that the $n$ hyper-parameter can be used to make trade-offs in network size, speed and accuracy. For example, if an accuracy score of approximately 89\% is acceptable on the QNLI task, then a modified ($n$ = 3) network could be used which would be approximately 1.5x smaller and 1.4x faster on GPUs. Other trade-offs can be made for situations where the network is to be deployed onto a system with very limited memory and computational resources. In that case, minimizing network size and computational complexity would be critical and choosing a modified ($n$ = $\infty$) network would decrease memory usage by more than 2x and be approximately 1.7x faster than the unmodified ($n$ = 1) network on GPU.

Additionally, we experiment on the BANDD and NoMLPLN techniques to understand their impact on fine-tuning accuracy. Although BANDD and NoMLPLN do not have an effect on the parameter count, they can improve speed by approximately 1.03x to 1.07x on different hardware platforms, while maintaining less than 2\% loss on most fine-tuning tasks. Interestingly, we notice that with NoMLPLN, the fine-tuning accuracy increases when compared to the unmodified network for certain tasks which may be due to over-fitting.

\begin{figure}[h]
    \center{\includegraphics[width=0.5\linewidth] {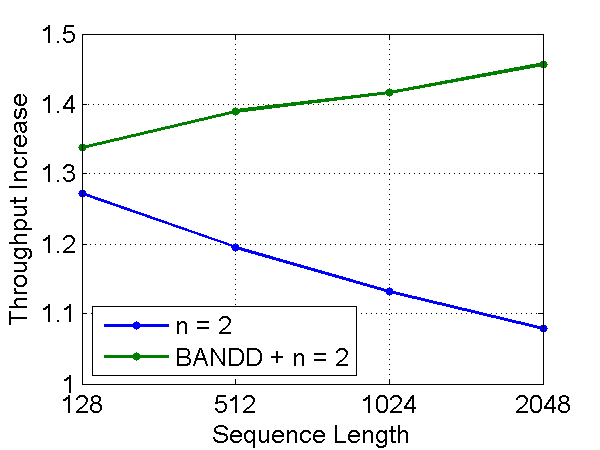}}
    \caption{\label{fig:bandd_n_2_throughput_increase} Throughput increase as a function of the sequence length for the $n$ = 2 and BAND + $n$ = 2 networks with batch size of 32 on a NVIDIA A100 GPU. Although the throughput increases for all sequence lengths for the $n$ = 2 variant, it further increases by also applying BANDD.}
\end{figure}

Figure \ref{fig:bandd_n_2_throughput_increase} shows the increase in throughput across different sequence lengths for the $n$ = 2 and BANDD + $n$ = 2 modified BERT\textsubscript{BASE} networks. From Figure \ref{fig:bandd_n_2_throughput_increase}, we see that BANDD better mitigates the \texttt{softmax} bottleneck with increasing sequence length. For example, at a sequence length of 1024, BANDD increases throughput to 1.41x from 1.13x with just $n$ = 2.

Finally, we investigate other complementary distillation techniques such as DistilBERT \cite{sanh2020distilbert}, BERT-PKD \cite{sun2019patient} and BERT-of-Theseus \cite{xu2020bertoftheseus} to show how network size and throughput compares to our approach. However, all of these techniques require the aid of a pre-trained teacher network. Table \ref{tab:glue_tasks} shows that some variants produced by our method are highly competitive with distillation techniques but without the need of an additional pass of training.

\section{Conclusion}

In this work we proposed different modifications to the BERT architecture to mitigate its complexity, which include reducing the number of intermediate layers and \texttt{layernorm} operations while also replacing all \texttt{softmax} operations. With these modifications to the BERT\textsubscript{BASE} network, we show a significant increase in throughput with a customizable trade-offs in fine-tuning accuracy.

For future work, we plan to apply our modifications to larger models like BERT\textsubscript{LARGE} and GPT to show that these changes can be used more widely. We also intend to integrate other network pruning techniques (e.g., ALBERT) to further decrease size and increase throughput.

\newpage
\appendix

\bibliographystyle{unsrt}  
\bibliography{acl,references}

\section{Appendix}
\label{sec:appendix}

\subsection{Code}
We used the open source BERT repository \url{https://github.com/NVIDIA/DeepLearningExamples/tree/master/TensorFlow/LanguageModeling/BERT} as the baseline to run pre-training experiments. We added our architectural modifications for intermediate layer removal, \texttt{softmax} replacement and \texttt{layernorm} removal to this repository locally. 

Additionally, we used the \url{ https://github.com/huggingface/transformers/tree/master/examples/pytorch/text-classification} repository to fine-tune our model variants on MNLI, QNLI SST-2 and QQP tasks, after transferring checkpoints from Tensorflow to PyTorch.

\subsection{Normalization Technique for Softmax Replacement}
The equations for the normalization technique proposed to replace \texttt{softmax} in \cite{richter2020normalized} are shown below:

\begin{equation}
\label{eq:equation_normalization}
    \textbf{a}_m^i = normalize([l_m^{i,1}, \dots, l_m^{i,N}]) \\
\end{equation}
where
\begin{equation}
\label{eq:equation_normalize}
  normalize(\textbf{x})^j = g\cdot\frac{x^j - \mu_\textbf{x}}{\sigma_\textbf{x}} + b
\end{equation}

In equation \ref{eq:equation_normalization}, $\textbf{l}_m$ denotes the attention logits vector obtained after the query-key dot product, $\textbf{a}_m$ represents the attention score vector for self-attention head $m$ and $N$ denotes the sequence length. In equation \ref{eq:equation_normalize}, $\mu_\textbf{x}$ and $\sigma_\textbf{x}$ are the mean and standard deviation of the input vector $\textbf{x}$  and $g$ and $b$ are learned gain and bias parameters, initialized to $1$ and $0$ respectively. We use this normalization technique in the BANDD reference unit shown in Figure \ref{fig:bandd_reference}.

\end{document}